\documentclass[runningheads]{llncs}
\usepackage{graphicx}
\usepackage{amsmath,amssymb} % define this before the line numbering.
\usepackage{color}
\usepackage[width=122mm,left=12mm,paperwidth=146mm,height=193mm,top=12mm,paperheight=217mm]{geometry}

\usepackage[table]{xcolor}
\usepackage{multirow}
\usepackage{tabularx}
\usepackage{booktabs}
\usepackage{hyperref}

\newcommand{\bim}[1]{\textcolor{white}{\fboxsep=0pt\fboxrule=2pt\fbox{\includegraphics[]{#1}}}}
\newcommand{\cim}[1]{\textcolor{yellow}{\fboxsep=0pt\fboxrule=2pt\fbox{\includegraphics[]{#1}}}}
\newcommand{\qim}[3]{\textcolor{white}{\fboxsep=0pt\fboxrule=0pt\fbox{
    \resizebox{#1\linewidth}{#2\linewidth}{\includegraphics[]{#3}}}
}}

\begin{document}
% \linenumbers
\pagestyle{headings}
\mainmatter
\def\ECCV18SubNumber{2990}  % Insert your submission number here

\title{Folded Recurrent Neural Networks for Future Video Prediction} % Replace with your title

\titlerunning{~}
\authorrunning{~}

\author{
Marc Oliu\\
Universitat Oberta de Catalunya\\
Centre de Visio per Computador\\
Rambla del Poblenou, 156, 08018 Barcelona\\
{\small moliusimon@uoc.edu}\\
\vspace{2mm}
Javier Selva\\
Universitat de Barcelona\\
Gran Via de les Corts Catalanes, 585, 08007 Barcelona\\
{\small javier.selva.castello@est.fib.upc.edu}\\
\vspace{2mm}
Sergio Escalera\\
Universitat de Barcelona\\
Centre de Visio per Computador\\
Gran Via de les Corts Catalanes, 585, 08007 Barcelona\\
{\small sergio@maia.ub.es}
}
\institute{}

\maketitle

% MOD Ponia "and"  AHORA OK MARC?????

\vspace{-3mm}
\begin{abstract} % 195 words
    %Main challenges in future video prediction are high variability in video content, temporal propagation of errors, and non-specificity of future frames: given a sequence of past frames there is a continuous distribution of possible futures. 
    Main challenges in future video prediction are high variability in videos, temporal propagation of errors, and non-specificity of future frames. This work introduces bijective Gated Recurrent Units (bGRU). Standard GRUs update a state, exposed as output, given an input. We extend them by considering the input as another recurrent state, and update it given the output using an extra set of logic gates. Stacking multiple such layers results in a recurrent auto-encoder: the operators updating the outputs comprise the encoder, while the ones updating the inputs form the decoder. Being the encoder and decoder states shared, the representation is stratified during learning: some information is not passed to the next layers. We show how only the encoder or decoder needs to be applied for encoding or prediction. This reduces the computational cost and avoids re-encoding predictions when generating multiple frames, mitigating error propagation. Furthermore, it is possible to remove layers from a trained model, giving an insight to the role of each layer. Our approach improves state of the art results on MMNIST and UCF101, being competitive on KTH with 2 and 3 times less memory usage and computational cost than the best scored approach. 
    \keywords{future video prediction, unsupervised learning, recurrent neural networks}
\end{abstract}

\section{Introduction}
\label{sec:introduction}

Future video prediction is a challenging task that recently received much attention due to its capabilities for learning in an unsupervised manner, making it possible to leverage large volumes of unlabelled data for video-related tasks such as action and gesture recognition \cite{srivastava2015unsupervised,lotter2016deep,liu2017video}, task planning \cite{oh2015action,ebert2017self}, weather prediction \cite{shi15nowcasting}, optical flow estimation \cite{patraucean2015spatio} and new view synthesis \cite{liu2017video}. 

One of the main problems in this task is the need of expensive models both in terms of memory and computational power in order to capture the variability present in video data. Another problem is the propagation of errors in recurrent models, which is tied to the inherent uncertainty of video prediction: given a series of previous frames, there are multiple feasible futures. This, left unchecked, results in a blurry prediction averaging the space of possible futures that propagates back into the network when predicting subsequent frames.

In this work we propose a new approach to recurrent auto-encoders (AE) with state sharing between encoder and decoder. We show how the exposed state in Gated Recurrent Units (GRU) can be used to create a bijective mapping between the input and output of each layer. To do so, the input is treated as a recurrent state, adding another set of logic gates to update it based on the output. Creating a stack of these layers allows for a bidirectional flow of information. Using the forward gates to encode inputs and the backward ones to generate predictions, we obtain a structure similar to an AE\footnote{Code available at \url{https://github.com/moliusimon/frnn}.}, but with many inherent advantages. It reduces memory and computational costs during both training and testing: only the encoder or decoder is executed for input encoding or prediction, respectively. Furthermore, the representation is stratified, encoding only part of the input at each layer: low level information not necessary to capture higher level dynamics is not passed to the next layer. Also, it naturally provides a noisy identity mapping of the input, facilitating the initial stages of training: the input to the first bGRU holds the last encoded frame or, if preceded by convolutional layers, an over-complete representation of the same. During generation, that first untrained bGRU randomly modifies the last input, introducing a noise signal. The approach also mitigates the propagation of errors: while it does not solve the problem of blur, it prevents its magnification in subsequent predictions. Moreover, a trained network can be deconstructed in order to analyse the role of each layer in the final predictions, making the model more explainable. Since the encoder and decoder states are shared, the architecture can be thought of as a recurrent AE folded in half, with encoder and decoder layers overlapping. We call our method Folded Recurrent Neural Network (fRNN). Our main contributions are: 1) A new shared-state recurrent AE with lower memory and computational costs. 2) Mitigation of error propagation through time. 3) It naturally provides an identity function during training. 4) Model explainability and optimisation through layer removal. 5) Demonstration of representation stratification.

%\begin{itemize}\small
%    \item A new shared-state recurrent AE with low memory and computational costs.
%    \item Mitigation of error propagation through time.
%    \item It naturally provides an identity function during training.
%    \item Model explainability and optimisation through layer removal.
%    \item Demonstration of representation stratification.
%\end{itemize}

%The rest of the paper is organised as follows. Section \ref{sec:sota} reviews related work. Section \ref{sec:method} presents the proposed method. Section \ref{sec:experiments} performs a quantitative and qualitative analysis of the method against state of the art alternatives. Finally, Section \ref{sec:conclusions} concludes the paper.

% --------------------------------------------------------
% --------------------------------------------------------
\section{Related work}
\label{sec:sota}

%----------------
% AE, CNN, LSTM
%----------------
While initial proposals focused on prediction on small patches \cite{ranzato2014baseline,michalski2014modeling}, future video prediction is nowadays generally approached by building a deep model capable of understanding the input sequence in a manner that allows for the generation of the following frames. 

\textbf{Building Blocks}. Due to the characteristics of the problem, an AE setting has been widely used \cite{srivastava2015unsupervised,oh2015action,finn2016unsupervised,villegas2017decomposing,denton2017disentangled}: the encoder extracts valuable information from the input and the decoder produces new frames. Generally, encoder and decoder are CNNs that tackle the spatial dimension. LSTMs are commonly used to handle the temporal dynamics and project the representations into the future. Some works compute the temporal dynamics at the deep representation bridging the encoder and decoder \cite{oh2015action,patraucean2015spatio,cricri2016video,denton2017disentangled}. Others jointly handle space and time by using Convolutional LSTMs \cite{finn2016unsupervised,lotter2016deep,patraucean2015spatio,kalchbrenner2017vpn,liang2017dual} (or GRUs, as in our case), which use convolutional kernels at their gates. For instance, Lotter \emph{et al.} \cite{lotter2016deep} use a recurrent residual network with convolutional LSTM where each layer minimises the discrepancies from previous block predictions. Common variations of the AE also include a conditional term to guide the temporal transform, such as a time differential \cite{vukotic2017one} or prior knowledge of scene events, reducing the space of possible futures. Oh \emph{et al.} \cite{oh2015action} predict future frames on Atari games conditioning on the action taken by the player. Some works propose such action conditioned models foreseeing an application for autonomous agents learning in an unsupervised fashion \cite{finn2016unsupervised,kalchbrenner2017vpn}. Finn \emph{et al.} \cite{finn2016unsupervised} predict a sequence of future frames within a physical system based on both previous frames and actions taken by a robotic arm interacting with the scene. The method was recently applied to task planning \cite{ebert2017self} and adapted to perform stochastic future frame prediction \cite{babaeizadeh2018stochastic}.

\textbf{Bridge connections}. Introducing bridge connections (connections between equivalent layers of encoder and decoder) is also common \cite{finn2016unsupervised,liu2017video,cricri2016video,villegas2017decomposing}. This allows for a stratified representation of the input sequence, reducing the capacity needs of subsequent layers. \emph{Video Ladder Networks} (VLN) \cite{cricri2016video} use a convolutional AE topology implementing skip connections. Pairs of convolutions are grouped into residual blocks, horizontally passing information between corresponding blocks, both by directly and by using a recurrent bridge layer. This topology was further extended with \emph{Recurrent Ladder Networks} (RLN) \cite{ilin2017recurrent}, where the recurrent bridge connections were removed, and the residual blocks replaced by recurrent layers. %These last take three inputs: cell state, output of the previous layer, and output of the corresponding layer from the previous encoding/decoding step. 
We propose an alternative to bridge connections by completely sharing the state between encoder and decoder, reducing computational needs while maintaining the stratification ability. 
% TODO cuando se comenta que no se pueden omitir el encoding/decoding steps también se podría mencionar que eso incrementará los errores con el tiempo
Both VLN and RLN share some similarities with our approach: they propose a recurrent AE with bridge connections between encoder and decoder. However, using skip connections instead of state sharing has some disadvantages: higher number of parameters and memory requirements, impossibility to skip the encoding/decoding steps (resulting in a higher computational cost) and reduced explainability due to not allowing layers to be removed after training. Finally, bridge connections do not provide an initial identity function during training. This makes it hard for the model to converge in some cases: when the background is homogeneous the model may not learn a proper initial mapping between input and output, but set the weights to zero and adjust the bias of the last layer, eliminating the gradient in the process.

\textbf{Prediction atom}. Most of the proposed architectures for future frame generation directly predict at pixel level (as in our case).
However, some models have been designed to predict motion and use it to transform the input into future frames. For instance, using the input sequence to anticipate optical flow \cite{liu2017video,patraucean2015spatio} or convolutional kernels \cite{brabandere2016dfn,xue2016visual}. Other methods propose mapping the input sequence onto predefined feature spaces, such as affine transforms \cite{amersfoort2017transformation} or human pose vectors \cite{walker2017pose}. These systems use sequences of such features instead of working directly at pixel level. Then, they use the predicted feature vectors to generate the next frames. 

%----------------
% Loss and GANs
%----------------
\textbf{Loss and GANs}. Commonly used loss functions such as L2 or MSE tend to average the space of possible futures. For this reason, some works\cite{mathieu2015deep,villegas2017decomposing,walker2017pose,liang2017dual} propose using Generative Adversarial Networks (GAN) \cite{goodfellow2014gan} to help traditional losses choose among possible futures by ensuring realistic looking frames and coherent sequences. Mathieu \emph{et al.} \cite{mathieu2015deep} use a plain multi-scale CNN in an adversarial setting and propose the Gradient Difference Loss to sharpen the predictions.

%----------------
% Disentangled Motion/Content
%----------------
\textbf{Disentangled Motion/Content}. Some authors encode content and motion separately. Villegas \emph{et al.} \cite{villegas2017decomposing} use an AE architecture with a two-stream encoder: for motion, a CNN + LSTM encodes difference images; for appearance, a plain CNN encodes the last input frame. %The decoder receives a concatenation of both and uses multi-scale residual connections.
In a similar fashion, Denton \emph{et al.} \cite{denton2017disentangled} use two separate encoders and an adversarial setting to obtain a disentangled representation of content and motion. %They trained separately an CNN AE for reconstruction and a LSTM to predict using sequences of encoded frames. 
Alternatively, some works predict motion and content in parallel to benefit from the combined strengths of both tasks. While Sedaghat \emph{et al.} \cite{sedaghat2016hybrid} propose using an encoding with a dual objective (flow and future frame), Liang \emph{et al.} \cite{liang2017dual} use a dual GAN setting and combine both predicted frame and motion to generate the actual next frame.

\textbf{Feedback Predictions}. Finally, an important aspect of the recurrent-based models is that they are based on the use of feedback predictions. In general, a model is trained to predict a specific number of time-steps into the future. In order to predict further in time they need to use their own predictions as input. This, if not handled properly, may accentuate small mistakes causing the predictions to quickly deteriorate over time. Our model solves this by enabling encoder and decoder to be executed any number of times independently. This is similar to the proposal by Srivastava \emph{et al.} \cite{srivastava2015unsupervised}, which uses a recurrent AE approach where an input sequence is encoded and its state copied into the decoder. The decoder is then applied to generate a given number of frames. However, it is limited to a single recurrent layer for each part.

%{\color{olive}Our model is based on a recurrent AE architecture to predict frames at pixel level. We use the proposed bijective GRUs with convolutional gates to handle both temporal and spatial dimensions. By sharing states between encoder and decoder instead of using bridge connections, the encoder states are updated every time a future frame is decoded. This avoids the need of feedback predictions.}

% --------------------------------------------------------
% --------------------------------------------------------
\section{Proposed method}
\label{sec:method}

We propose an architecture based on recurrent convolutional AEs to deal with the network capacity and error propagation problems for future video prediction. It consists on a series of bijective GRU layers, which allow for a bidirectional flow of information between input and output: they consider the input as a recurrent state and update it using an extra set of gates. These are then stacked, forming an encoder and decoder using, respectively, the forward and backward functions of the bijective GRUs (Fig.\ref{fig:topology}). We call it Folded Recurrent Neural Network (fRNN). Because of the state sharing between encoder and decoder, the topology allows for: stratification of the encoded information, lower memory and computational requirements compared to regular recurrent AEs, mitigated propagation of errors, and increased explainability through layer removal.

% --------------------------------------------------------
\subsection{Bijective Gated Recurrent Units}
\label{sec:method:bjgru}

\begin{figure}[t!]
    \centering
    \includegraphics[width=11.4cm]{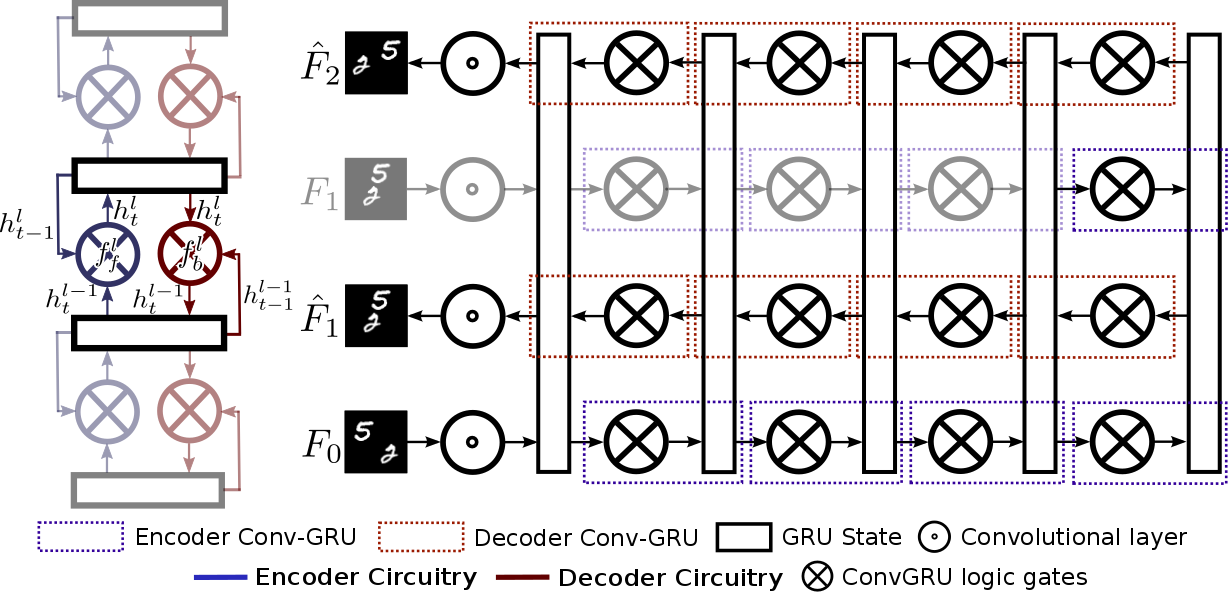}
   \caption{\textbf{Left:} Scheme of a bGRU. Shadowed areas illustrate how multiple bGRU layers are stacked. \textbf{Right:} fRNN topology. The recurrent states of the encoder and decoder are shared, resulting in  a bidirectional mapping between states. Shadowed areas represent unnecessary circuitry: re-encoding of the predictions is avoided thanks to the decoder updating all the states. \textbf{Left-Right:} Blue and red correspond to forward and backward gates, respectively. Rectangles represent the recurrent state cell.}\vspace{-0.4cm}
    \label{fig:topology}
\end{figure}

GRUs have their state fully exposed as output. This allows us to define a bidirectional mapping between input and output by replicating the logic gates of the GRU layer. To do so, we consider the input as a state on itself. Lets define the output of a GRU at layer $l$ and time step $t$ as $h^{l}_{t} = f^{l}_{f}(h^{l-1}_{t}, h^{l}_{t-1})$ given an input $h^{l-1}_{t}$ and its state at the previous time step $h^{l}_{t-1}$. A second set of weights can be used to define an inverse mapping $h^{l-1}_{t} = f^{l}_{b}(h^{l}_{t}, h^{l-1}_{t-1})$ using the output of the forward function at the current time step to update its input, which is treated as the hidden state of the inverse function. This is illustrated in Fig. \ref{fig:topology}. We will refer to this double mapping as bijective GRU (bGRU).

% --------------------------------------------------------
\subsection{Folded Recurrent Neural Network}
\label{sec:method:frnn}

By stacking multiple bGRUs, a recurrent AE is obtained. Given $n$ bGRUs, the encoder is defined by the set of forward functions $E = \{f^{1}_{f},~...,~f^{n}_{f}\}$ and the decoder by the set of backward functions $D = \{f^{n}_{b},~...,~f^{1}_{b}\}$. This is illustrated in Fig. \ref{fig:topology}, and is equivalent to a recurrent AE, but with shared states, having 3 main advantages: 1) It is not necessary to feed the predictions back into the network in order to generate the following predictions. Because the states are shared, the decoder already updates all the states except for the bridge state between encoder and decoder. The bridge state is updated by applying the last layer of the encoder before generating the next prediction. The shadowed area in Fig. \ref{fig:topology} shows the section of the computational graph that is not required when performing multiple sequential predictions. For the same reason, when considering multiple sequential elements before prediction, only the encoder is required. 2) Because the network updates its states from the higher level representations to the lowest ones during prediction, errors introduced at a given layer during generation are not propagated back into deeper layers, leaving the higher-level dynamics unaffected. 3) The model implicitly provides a noisy identity model during training, as it is shown in Fig. \ref{fig:stratified:mmnist}, when all bGRU layers are removed. The input state of the first bGRU layer is either the input image itself or, when first applying a series of convolutional layers, an over-complete representation of the input. A noise signal is then introduced to the representation by the backward function of the untrained first bGRU layer. Consequently providing the model with an initial identity model. As we show in Section \ref{sec:experiments:quantitative}, this helps the model to converge in some datasets like MMNIST: when the same background is shared across instances, it prevents the model from killing the gradients by adjusting the biases to match the background and setting the weights to zero.

This approach shares some similarities with VLN and RLN. As with them, part of the information can be passed directly between corresponding layers of the encoder and decoder, not having to encode a full representation of the input into the deepest layer. However, our model implicitly passes the information through the shared recurrent states, making bridge connections unnecessary. When compared against an equivalent recurrent AE with bridge connections, this results in a much lower computational and memory cost. More specifically, the number of weights in a pair of forward and backward functions is equal to $3(\overline{h^{l-1}}^{2} + \overline{h^{l}}^{2} + 2\overline{h^{l-1}}~\overline{h^{l}})$ in the case of bGRU, where $\overline{h^{l}}$ corresponds to the state size of layer $l$. When using bridge connections, as in the case of VLN and RLN, that value is incremented to $3(\overline{h^{l-1}}^{2} + \overline{h^{l}}^{2} + 4\overline{h^{l-1}}~\overline{h^{l}})$. This corresponds to an increase of 44\% in the number of parameters when one state has double the size of the other, and of 50\% when they have the same size. Furthermore, both the encoder and decoder must be applied at each time step. Thus, memory usage is doubled and computational cost is increased by a factor of between $2.88$ and $3$.

% --------------------------------------------------------
\subsection{Training Folded RNNs}
\label{sec:method:training}

% MOD Ponía "folded RNN"
In a regular recurrent AE, a ground truth frame is introduced at each time step by applying both encoder and decoder. The output is used as a supervision point, comparing it to the next ground truth frame in the sequence. This implies all predictions are at a single time step from the last ground truth prediction. Here, we propose a training approach for fRNNs that exploits the ability of the topology of skipping the model encoder or decoder at a given time step. First $g$ ground truth frames are shown to the network by passing them through the encoder. The decoder is then applied $p$ times, producing $p$ predictions. This results in only half the memory requirements: either encoder or decoder is applied at each step, never both. This has the same advantage as the approach by Srivastava \cite{srivastava2015unsupervised}, where recurrently applying the decoder without further ground truth inputs encourages the network to learn video dynamics. This also prevents the network from learning an identity model, i.e. copying the last input to the output.

% --------------------------------------------------------
% --------------------------------------------------------
\section{Experiments}
\label{sec:experiments}

Here, we first discuss data, evaluation protocol, and methods. Then we provide a detailed quantitative and qualitative evaluation. We finish with a brief analysis on the stratification of sequence representation among bGRU layers\footnote{Additional qualitative results, as well as an extended table of quantitative results by time step, are shown in the supplementary material.}.

%We analyse our proposal on three public datasets and compare its performance against state of the art alternatives. 
%In Section \ref{sec:experiments:methodology} we introduce the considered data and evaluation protocol, and in Section \ref{sec:experiments:methods} we present the methods used for comparison. Quantitative and qualitative analyses are presented in Sections \ref{sec:experiments:quantitative} and \ref{sec:experiments:qualitative}, respectively. Finally, an analysis of the representation stratification of the proposed topology is performed in Section \ref{sec:experiments:stratification}.

% --------------------------------------------------------
\subsection{Data and evaluation protocol}
\label{sec:experiments:methodology}

We considered 3 datasets of different complexity in order to analyse the performance of the proposed method: Moving MNIST (MMNIST)\cite{srivastava2015unsupervised}, KTH \cite{kth}, and UCF101 \cite{ucf101}. MMNIST consists of $64\times64$ grayscale sequences of length 20 displaying pairs of digits moving around the image. The sequences are generated by randomly sampling pairs of digits and trajectories. It contains a fixed test partition with 10000 sequences. We generated a million extra samples for training. KTH consists of 600 videos of 15-20 seconds with $25$ subjects performing 6 actions in 4 different settings. The videos are grayscale, at a resolution of $120\times160$ pixels and 25 fps. The dataset has been split into subjects 1 to 16 for training, and 17 to 25 for testing, resulting in 383 and 216 sequences, respectively. Frame size is reduced to $64\times80$ using bilinear interpolation, removing 5 pixels from the left and right borders before resizing. UCF101 displays 101 actions, such as playing instruments, weight lifting or sports. It is the most challenging dataset considered, with a high intra-class variability. It contains 9950 training sequences and 3361 test sequences. These are RGB at a resolution of $320\times240$ pixels and 25 fps. To increase motion between consecutive frames, one of every two frames was removed. Following the same procedure as with KTH, the frame size is reduced to $64\times85$.

All methods are tested using 10 input frames to generate the following 10 frames. We use 3 common metrics for video prediction analysis: Mean Squared Error (MSE), Peak Signal-to-Noise Ratio (PSNR), and Structural Dissimilarity (DSSIM). MSE and PSNR are objective measurements of reconstruction quality. DSSIM is a measure of the perceived quality. For DSSIM we use a Gaussian sliding window of size $11\times11$ and $\sigma=1.5$.

% --------------------------------------------------------
\subsection{Methods}
\label{sec:experiments:methods}

To train the proposed method we used RMSProp with learning rate of 0.0001 and batch size of 12, sampling a random sub-sequence at each epoch. Weights were orthogonally initialised, with biases set to 0. For testing, we considered all sub-sequences of length 20. Our network topology consists of two convolutional layers followed by 8 convolutional bGRU layers, applying a $2\times2$ max pooling every 2 layers. Topology details are shown in Table \ref{fig:topology:details}. We use deconvolution and nearest neighbours interpolation to invert the convolutional and max pooling layers, respectively. We train with L1 loss.

\begin{table}[t!]
    \centering
    \resizebox{\linewidth}{!}{\begin{tabular}{r|c|c|c|c|c|c|c|c|c|c|c|c|c|c|}
         \multicolumn{1}{c}{} & \multicolumn{1}{c}{Conv 1} & \multicolumn{1}{c}{Conv 2} & \multicolumn{1}{c}{Pool 1} & \multicolumn{1}{c}{bGRU 1} & \multicolumn{1}{c}{bGRU 2} & \multicolumn{1}{c}{Pool 2} & \multicolumn{1}{c}{bGRU 3} & \multicolumn{1}{c}{bGRU 4} & \multicolumn{1}{c}{Pool 3} & \multicolumn{1}{c}{bGRU 5} & \multicolumn{1}{c}{bGRU 6} & \multicolumn{1}{c}{Pool 4} & \multicolumn{1}{c}{bGRU 7} & \multicolumn{1}{c}{bGRU 8} \\ \cline{2-15}
         Num. Units  & 32 & 64 & - & 128 & 128 & - & 256 & 256 & - & 512 & 512 & - & 256 & 256 \\ \cline{2-15}
         Kernel size & $5\times5$ & $5\times5$ & $2\times2$ & $5\times5$ & $5\times5$ & $2\times2$ & $5\times5$ & $5\times5$ & $2\times2$ & $3\times3$ & $3\times3$ & $2\times2$ & $3\times3$ & $3\times3$ \\ \cline{2-15}
         Stride      & 1 & 1 & 2 & 1 & 1 & 2 & 1 & 1 & 2 & 1 & 1 & 2 & 1 & 1 \\ \cline{2-15}
         Activation  & tanh & tanh & - & \multicolumn{2}{c|}{sigmoid \& tanh} & - & \multicolumn{2}{c|}{sigmoid \& tanh} & - & \multicolumn{2}{c|}{sigmoid \& tanh} & - & \multicolumn{2}{c|}{sigmoid \& tanh} \\ \cline{2-15}
    \end{tabular}}
    \vspace{0.5mm}
    \caption{Parameters of the topology used for the experiments. The decoder applies the same topology in reverse, using nearest neighbours interpolation and  transposed convolutions to revert the pooling and convolutional layers.}
    \label{fig:topology:details}\vspace{-0.5cm}
\end{table}

For evaluation, we include a stub baseline model predicting the last input frame, and design a second baseline (RLadder) to evaluate the advantages of using state sharing. RLadder has the same topology as the fRNN model, but uses bridge connections instead of state sharing. Note that to keep the same state size on GRU layers, using bridge connections doubles the memory size and almost triples the computational cost (Sec.\ref{sec:method:frnn}). This is similar to how RLN \cite{ilin2017recurrent} works, but using regular conv GRU layers in the decoder. We also compare against Srivastava \cite{srivastava2015unsupervised} and Mathieu \cite{mathieu2015deep}. The former only handles the temporal dimension explicitly with LSTMs, while the latter only treats the spatial dimensions using 3D CNN. Next, we compare against Villegas \cite{villegas2017decomposing}, which, contrary to our proposal, uses feedback predictions. 
%Furthermore, and opposed to the baselines, they handle separately time and space and use a more complex loss combining L2, GDL and an adversarial setting while we only use L1. 
Finally, we compare against Lotter \emph{et al.} \cite{lotter2016deep} which is based on residual error reduction. All of them were adapted to train using 10 frames as input and predicting the next 10, using the topologies and parameters defined by the authors.

% --------------------------------------------------------
\subsection{Quantitative analysis}
\label{sec:experiments:quantitative}

\begin{figure}[t!]
    \centering
    \vspace{2mm}
    \includegraphics[width=\linewidth]{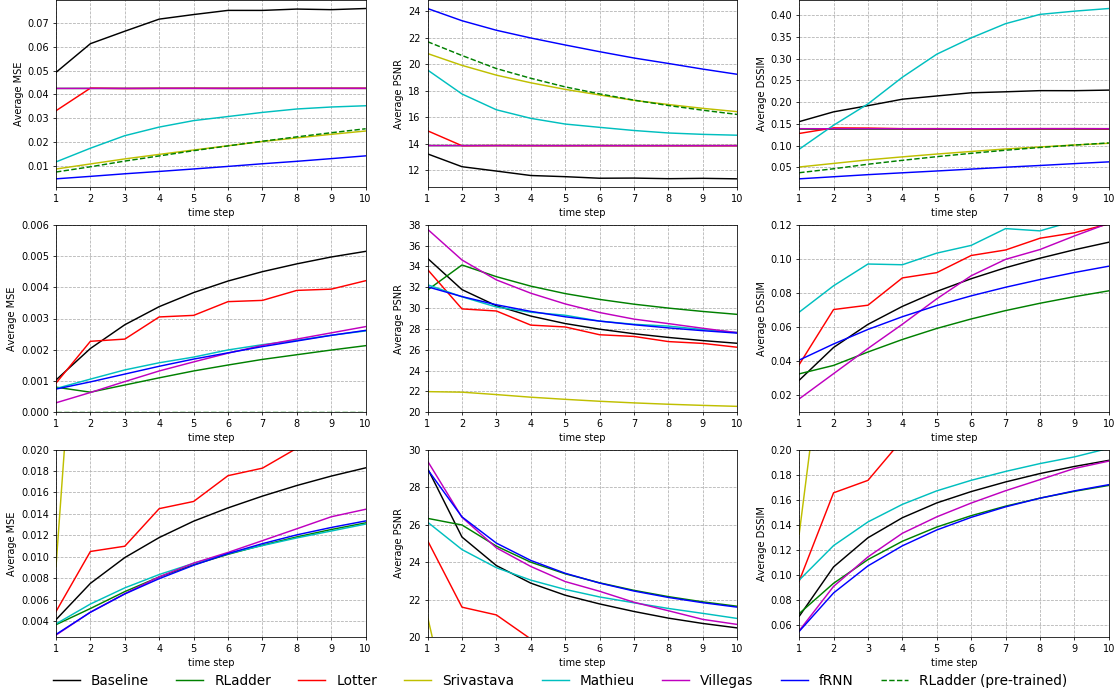}
    \vspace{-8mm}
    \caption{Quantitative results on the considered datasets in terms of the number of time steps since the last input frame. From top to bottom: MMNIST, KTH, and UCF101. From left to right: MSE, PSNR, and DSSIM. For MMNIST, RLadder is pre-trained to learn an initial identity mapping, allowing it to converge.}
    \label{fig:quantitative}
\end{figure}

The first row of Fig. \ref{fig:quantitative} displays the results for the MMNIST dataset for the proposed method, baselines, and state of the art alternatives. Mean scores are shown in Table \ref{tab:results}. fRNN performs best on all time steps and metrics, followed by Srivastava \emph{et al.} \cite{srivastava2015unsupervised}. These two are the only methods to provide valid predictions on this dataset. Most other methods predict a black frame, with Mathieu \emph{et al.} \cite{mathieu2015deep} progressively blurring the digits. This is caused by a loss of gradient during the first stages of training. On more complex datasets the methods start by learning an identity function, then refining the results. This is possible since in many sequences most of the frame remains unchanged. In the case of MMNIST, where the background is homogeneous, it is much easier for the models to set the weights of the output layer to zero and set the biases to match the background colour. This cuts the gradient and prevents further learning. Srivastava \emph{et al.} \cite{srivastava2015unsupervised} use an auxiliary decoder to reconstruct the input frames, forcing the model to learn an identity function. This, as discussed at the end of Section \ref{sec:method:frnn}, is implicitly handled in our method, giving an initial solution to improve on and preventing the models from learning a black image. In order to verify this effect, we pre-trained RLadder on the KTH dataset. While this dataset has completely different dynamics, the initial step to solve the problem remains: providing an identity function. Afterwards the model is fine-tuned on the MMNIST dataset. As it is shown in Fig. \ref{fig:quantitative} (dashed lines), this results in the model converging, with an accuracy comparable to Srivastava \emph{et al. \cite{srivastava2015unsupervised}} for the 3 evaluation metrics.

\begin{table}[t!]
    \centering
    \resizebox{\linewidth}{!}{
    \begin{tabular}{r|c|c|c|}
        \multicolumn{1}{c}{} & \multicolumn{3}{c}{MMNIST} \\ \cline{2-4}
                                                     & MSE     & PSNR   & DSSIM \\ \cline{2-4}
        Baseline                                     & 0.06989 & 11.745 & 0.20718 \\ \cline{2-4}
        RLadder                                      & 0.04254 & 13.857 & 0.13788 \\ \cline{2-4}
        %Prednet \cite{lotter2016deep}                & 0.04137 & 14.017 & 0.14201 \\ \cline{2-4}
        Lotter \cite{lotter2016deep}                & 0.04161 & 13.968 & 0.13825 \\ \cline{2-4}
        Srivastava \cite{srivastava2015unsupervised} & 0.01737 & 18.183 & 0.08164 \\ \cline{2-4}
        %Mathieu \cite{mathieu2015deep}               & 0.03071 & 15.361 & 0.32770 \\ \cline{2-4}
        Mathieu \cite{mathieu2015deep}               & 0.02748 & 15.969 & 0.29565 \\ \cline{2-4}
        Villegas \cite{villegas2017decomposing}      & 0.04254 & 13.857 & 0.13896 \\ \cline{2-4}
        fRNN                                         & \textbf{0.00947} & \textbf{21.386} & \textbf{0.04376} \\ \cline{2-4}
    \end{tabular}
    \begin{tabular}{|c|c|c|}
        \multicolumn{3}{c}{KTH} \\ \cline{1-3}
        MSE     & PSNR   & DSSIM \\ \cline{1-3}
        0.00366 & 29.071 & 0.07900 \\ \cline{1-3}
        \textbf{0.00139} & \textbf{31.268} & \textbf{0.05945} \\ \cline{1-3}
        %0.00807 & 24.635 & 0.13588 \\ \cline{1-3}
        0.00309 & 28.424 & 0.09170 \\ \cline{1-3}
        0.00995 & 21.220 & 0.19860 \\ \cline{1-3} 
        %0.00194 & 29.097 & 0.10018 \\ \cline{1-3}
        0.00180 & 29.341 & 0.10410 \\ \cline{1-3}
        0.00165 & 30.946 & 0.07657 \\ \cline{1-3}
        0.00175 & 29.299 & 0.07251 \\ \cline{1-3}
    \end{tabular}
    \begin{tabular}{|c|c|c|}
        \multicolumn{3}{c}{UCF101} \\ \cline{1-3}
        MSE     & PSNR   & DSSIM   \\ \cline{1-3}
        0.01294 & 22.859 & 0.15043 \\ \cline{1-3}
        0.00918 & 23.558 & 0.13395 \\ \cline{1-3}
        %0.02124 & 20.398 & 0.19013 \\ \cline{1-3}
        0.01550 & 19.869 & 0.21389 \\ \cline{1-3}
        0.14866 & 10.021 & 0.42555 \\ \cline{1-3}
        %0.01287 & 20.492 & 0.20730 \\ \cline{1-3}
        0.00926 & 22.781 & 0.16262 \\ \cline{1-3}
        0.00940 & 23.457 & 0.14150 \\ \cline{1-3}
        \textbf{0.00908} & \textbf{23.872} & \textbf{0.13055} \\ \cline{1-3}
    \end{tabular}}
    \caption{Average results over 10 time steps.}
    \vspace{-0.7cm}
    \label{tab:results}
\end{table}

On the KTH dataset, Table \ref{tab:results} shows the best approach is our RLadder baseline followed by fRNN and Villegas \emph{et al.} \cite{villegas2017decomposing}, both having similar results, but with Villegas \emph{et al.} having slightly lower MSE and higher PSNR, and fRNN a lower DSSIM. While both approaches obtain comparable average results, the error increases faster over time in the case of Villegas \emph{et al.} (second row in Fig.\ref{fig:quantitative}). Mathieu obtains good scores for MSE and PSNR, but has a much worse DSSIM.

For the UCF101 dataset, as shown in Table \ref{tab:results}, our fRNN approach is the best performing for all 3 metrics. When looking at the third row of Fig. \ref{fig:qualitative:ucf101}, one can see that Villegas \emph{et al.} starts out with results similar to fRNN on the first frame, but as in the case of KTH and MMNIST, the predictions degrade faster than with the proposed approach. Two methods display low performance in most cases. Lotter \emph{et al.} works well for the first predicted frame in the case of KTH and UCF101, but the error rapidly increases on the following predictions. This is due to a magnification of artefacts introduced on the first prediction, making the method unable to predict multiple frames without supervision. In the case of Srivastava \emph{et al.} the problem is about capacity: it uses fully connected LSTM layers, making the number of parameters explode quickly with the state cell size. This severely limits the representation capacity for complex datasets such as KTH and UCF101.

Overall, for the considered methods, fRNN is the best performing on MMINST and UCF101, the later being the most complex of the 3 datasets. We achieved these results with a simple topology: apart from the proposed bGRU layers, we use conventional max pooling with an L1 loss. There are no normalisation or regularisation mechanisms, specialised activation functions, complex topologies or image transform operators. In the case of MMNIST, fRNN shows the ability to find a good initial representation and converges to good predictions where most other methods fail. In the case of KTH, fRNN has an overall accuracy comparable to that of Villegas \emph{et al.}, being more stable over time. It is only surpassed by the proposed RLadder baseline, a method equivalent to fRNN but with 2 and 3 times more memory and computational requirements.

% --------------------------------------------------------
\subsection{Qualitative analysis}
\label{sec:experiments:qualitative}

%In this section we evaluate our approach qualitatively, observing how the model behaves in terms of predicting movement dynamics, reconstructing occluded regions and specially on the propagation of blur. We then compare predictions from our approach against those of other state of the art methods.

\begin{figure}[t!]
    \centering
    \setlength{\tabcolsep}{0pt}
    \resizebox{\columnwidth}{!}{\begin{tabular}{ccccccccccccccc}
        \bim{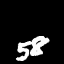} & 
        \bim{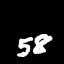} & 
        \bim{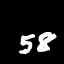} & 
        \bim{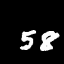} & 
        \bim{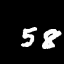} & 
        \bim{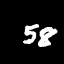} & 
        \bim{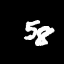} & 
        \bim{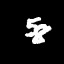} & 
        \bim{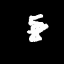} & 
        \bim{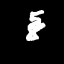} & 
        \bim{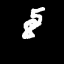} & 
        \bim{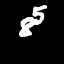} & 
        \bim{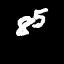} & 
        \bim{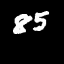} & 
        \bim{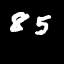} \\ 
        & & & & &
        \cim{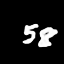} & 
        \cim{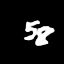} & 
        \cim{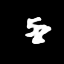} & 
        \cim{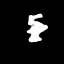} & 
        \cim{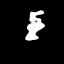} & 
        \cim{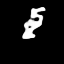} & 
        \cim{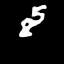} & 
        \cim{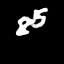} & 
        \cim{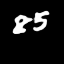} & 
        \cim{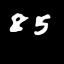} \\
        
        \bim{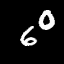} & 
        \bim{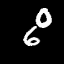} & 
        \bim{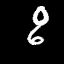} & 
        \bim{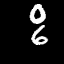} & 
        \bim{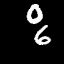} & 
        \bim{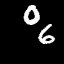} & 
        \bim{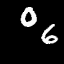} & 
        \bim{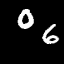} & 
        \bim{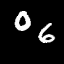} & 
        \bim{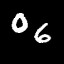} & 
        \bim{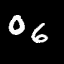} & 
        \bim{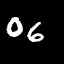} & 
        \bim{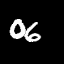} & 
        \bim{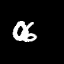} & 
        \bim{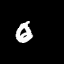} \\ 
        & & & & &
        \cim{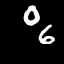} & 
        \cim{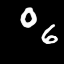} & 
        \cim{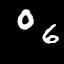} & 
        \cim{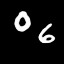} & 
        \cim{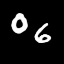} & 
        \cim{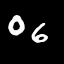} & 
        \cim{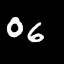} & 
        \cim{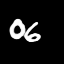} & 
        \cim{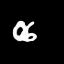} & 
        \cim{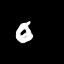} \\
                
        \bim{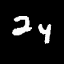} & 
        \bim{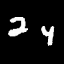} & 
        \bim{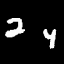} & 
        \bim{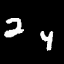} & 
        \bim{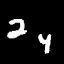} & 
        \bim{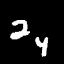} & 
        \bim{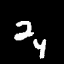} & 
        \bim{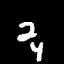} & 
        \bim{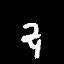} & 
        \bim{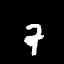} & 
        \bim{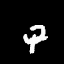} & 
        \bim{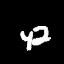} & 
        \bim{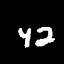} & 
        \bim{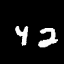} & 
        \bim{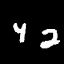} \\ 
        & & & & &
        \cim{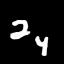} & 
        \cim{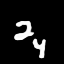} & 
        \cim{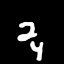} & 
        \cim{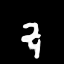} & 
        \cim{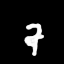} & 
        \cim{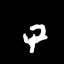} & 
        \cim{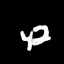} & 
        \cim{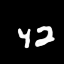} & 
        \cim{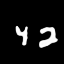} & 
        \cim{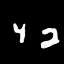} \\
    \end{tabular}}
    \vspace{-3mm}
    \caption{fRNN predictions on MMNIST. First row for each sequence shows last 5 inputs and target frames. Yellow frames are model predictions.}
    \label{fig:qualitative:mmnist}\vspace{-0.5cm}
\end{figure}

In this section we evaluate our approach qualitatively on some samples from the three considered datasets. Fig. \ref{fig:qualitative:mmnist} shows the last 5 input frames from some MMNIST sequences along with the next 10 ground truth frames and their corresponding fRNN predictions. The 10 predictions are generated sequentially without showing the previous ground truth/prediction to the network, that is, only using the decoder. As it can be seen, the digits maintain their sharpness across the sequence of predictions. Also, the bounces at the edges of the image are done correctly, and the digits do not distort or deform when crossing. This shows the network internally encodes the appearance of each digit, making it possible to reconstruct them after sharing the same region in the image plane.

\begin{figure}[t!]
    \centering
    \setlength{\tabcolsep}{0pt}
    \resizebox{\textwidth}{!}{\begin{tabular}{ccccccccccccccc}
        \bim{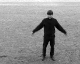} & 
        \bim{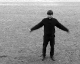} & 
        \bim{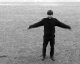} & 
        \bim{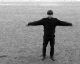} & 
        \bim{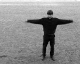} & 
        \bim{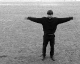} & 
        \bim{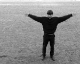} & 
        \bim{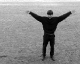} & 
        \bim{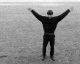} & 
        \bim{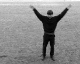} & 
        \bim{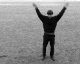} & 
        \bim{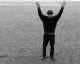} & 
        \bim{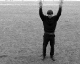} & 
        \bim{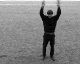} & 
        \bim{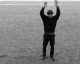} \\ 
        & & & & &
        \cim{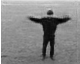} & 
        \cim{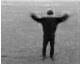} & 
        \cim{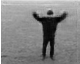} & 
        \cim{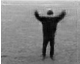} & 
        \cim{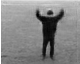} & 
        \cim{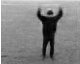} & 
        \cim{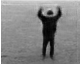} & 
        \cim{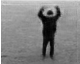} & 
        \cim{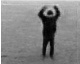} & 
        \cim{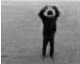} \\
        
        \bim{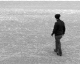} & 
        \bim{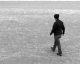} & 
        \bim{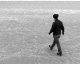} & 
        \bim{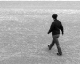} & 
        \bim{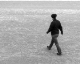} & 
        \bim{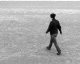} & 
        \bim{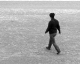} & 
        \bim{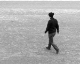} & 
        \bim{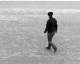} & 
        \bim{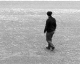} & 
        \bim{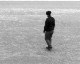} & 
        \bim{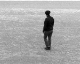} & 
        \bim{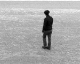} & 
        \bim{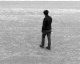} & 
        \bim{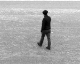} \\ 
        & & & & &
        \cim{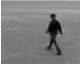} & 
        \cim{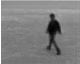} & 
        \cim{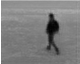} & 
        \cim{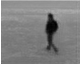} & 
        \cim{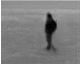} & 
        \cim{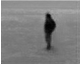} & 
        \cim{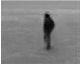} & 
        \cim{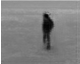} & 
        \cim{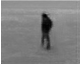} & 
        \cim{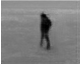} \\
        
        \bim{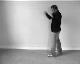} & 
        \bim{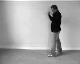} & 
        \bim{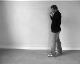} & 
        \bim{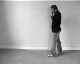} & 
        \bim{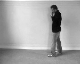} & 
        \bim{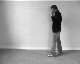} & 
        \bim{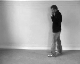} & 
        \bim{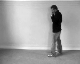} & 
        \bim{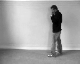} & 
        \bim{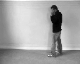} & 
        \bim{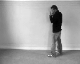} & 
        \bim{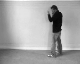} & 
        \bim{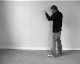} & 
        \bim{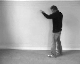} & 
        \bim{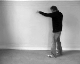} \\ 
        & & & & &
        \cim{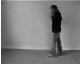} & 
        \cim{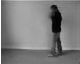} & 
        \cim{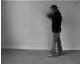} & 
        \cim{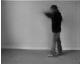} & 
        \cim{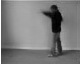} & 
        \cim{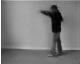} & 
        \cim{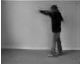} & 
        \cim{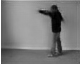} & 
        \cim{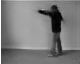} & 
        \cim{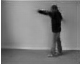} \\
    \end{tabular}}
    \vspace{-4mm}
    \caption{fRNN predictions on KTH. First row for each sequence shows last 5 inputs and target frames. Yellow frames are model predictions.}
    \label{fig:qualitative:kth}
\end{figure}

Qualitative examples of fRNN predictions on the KTH dataset are shown in Fig. \ref{fig:qualitative:kth}. It shows three actions: hand waving, walking, and boxing. The blur stops increasing after the first three predictions, generating plausible motions for the corresponding actions while background artefacts are not introduced. Although the movement patterns for each type of action have a wide range of variability on its trajectory, bGRU gives relatively sharp predictions for the limbs. The first and third examples also show the ability of the model to recover from blur. The blur slightly increases for the arms while the action is performed, but decreases again as these reach the final position.

\begin{figure}[t!]
    \centering
    \setlength{\tabcolsep}{0pt}
    \resizebox{\textwidth}{!}{\begin{tabular}{ccccccccccccccc}
        \bim{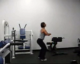} & 
        \bim{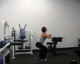} & 
        \bim{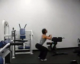} & 
        \bim{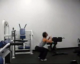} & 
        \bim{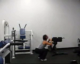} & 
        \bim{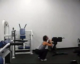} & 
        \bim{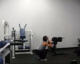} & 
        \bim{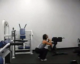} & 
        \bim{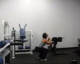} & 
        \bim{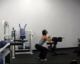} & 
        \bim{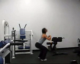} & 
        \bim{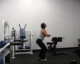} & 
        \bim{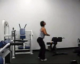} & 
        \bim{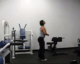} & 
        \bim{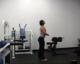} \\ 
        & & & & &
        \cim{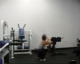} & 
        \cim{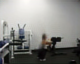} & 
        \cim{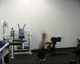} & 
        \cim{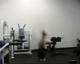} & 
        \cim{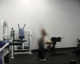} & 
        \cim{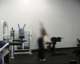} & 
        \cim{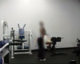} & 
        \cim{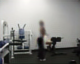} & 
        \cim{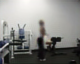} & 
        \cim{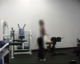} \\
        
        \bim{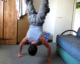} & 
        \bim{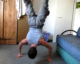} & 
        \bim{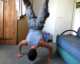} & 
        \bim{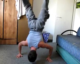} & 
        \bim{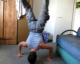} & 
        \bim{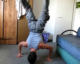} & 
        \bim{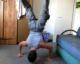} & 
        \bim{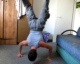} & 
        \bim{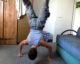} & 
        \bim{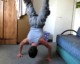} & 
        \bim{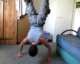} & 
        \bim{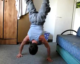} & 
        \bim{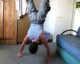} & 
        \bim{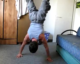} & 
        \bim{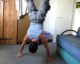} \\ 
        & & & & &
        \cim{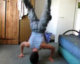} & 
        \cim{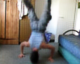} & 
        \cim{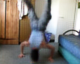} & 
        \cim{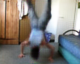} & 
        \cim{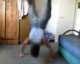} & 
        \cim{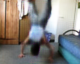} & 
        \cim{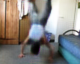} & 
        \cim{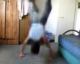} & 
        \cim{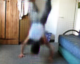} & 
        \cim{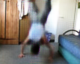} \\
        
        \bim{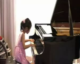} & 
        \bim{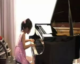} & 
        \bim{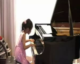} & 
        \bim{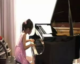} & 
        \bim{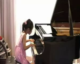} & 
        \bim{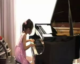} & 
        \bim{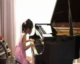} & 
        \bim{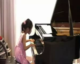} & 
        \bim{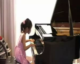} & 
        \bim{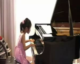} & 
        \bim{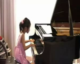} & 
        \bim{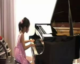} & 
        \bim{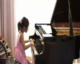} & 
        \bim{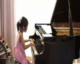} & 
        \bim{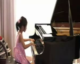} \\ 
        & & & & &
        \cim{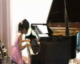} & 
        \cim{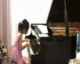} & 
        \cim{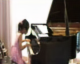} & 
        \cim{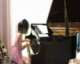} & 
        \cim{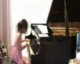} & 
        \cim{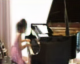} & 
        \cim{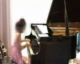} & 
        \cim{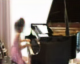} & 
        \cim{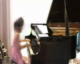} & 
        \cim{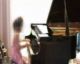} \\
    \end{tabular}}
    \vspace{-4mm}
    \caption{fRNN predictions on UCF. First row for each sequence shows last 5 inputs and target frames. Yellow frames are model predictions.}
    \vspace{-0.2cm}
    \label{fig:qualitative:ucf101}
\end{figure}

Fig. \ref{fig:qualitative:ucf101} shows fRNN predictions on the UCF101 dataset. These correspond to two different physical exercises and a girl playing the piano. Common to all predictions, the static parts do not lose sharpness over time, and the background is properly reconstructed after an occlusion. The network correctly predicts actions with low variability, as shown in rows 1-2, where a repetitive movement is performed, and in last row, where the girl recovers a correct body posture. These dynamic regions introduce blur due to the uncertainty of the action, averaging the possible futures. The first row also shows an interesting behaviour: while the woman is standing up, the upper body becomes blurry due to uncertainty, but as the woman finishes her motion and ends up in the expected upright position, the frames sharpen again. Since the model does not propagate errors to deeper layers nor makes use of previous predictions for the following ones, the introduction of blur does not imply this blur will be propagated. In this example, while the middle motion could have multiple predictions depending on the movement pace and the inclination of the body while performing it, the final body pose has a lower uncertainty.

\begin{figure}[t!]
    \centering
    \setlength{\tabcolsep}{0pt}
    \resizebox{\linewidth}{!}{\begin{tabular}{ccc@{\hspace{1.5mm}}ccc@{\hspace{1.5mm}}ccc}
        \Large{Input 1} & \Large{Input 5} & \Large{Input 10} & \Large{GT 1} & \Large{GT 5} & \Large{GT 10} & & & \\
        \addlinespace[0.5mm]
        \qim{0.265}{0.265}{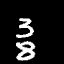} &
        \qim{0.265}{0.265}{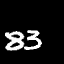} &
        \qim{0.265}{0.265}{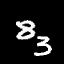} &
        \qim{0.265}{0.265}{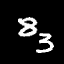} &
        \qim{0.265}{0.265}{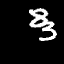} &
        \qim{0.265}{0.265}{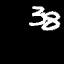} \\
        \qim{0.265}{0.20}{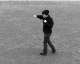} &
        \qim{0.265}{0.20}{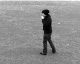} &
        \qim{0.265}{0.20}{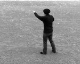} &
        \qim{0.265}{0.20}{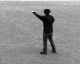} &
        \qim{0.265}{0.20}{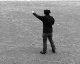} &
        \qim{0.265}{0.20}{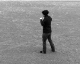} \\
        \qim{0.265}{0.20}{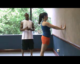} &
        \qim{0.265}{0.20}{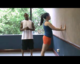} &
        \qim{0.265}{0.20}{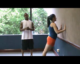} &
        \qim{0.265}{0.20}{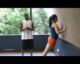} &
        \qim{0.265}{0.20}{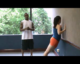} &
        \qim{0.265}{0.20}{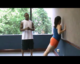} & & & \\
        
        \addlinespace[5mm]
        \Large{Pred. 1} & \Large{Pred. 5} & \Large{Pred. 10} & \Large{Pred. 1} & \Large{Pred. 5} & \Large{Pred. 10} & \Large{Pred. 1} & \Large{Pred. 5} & \Large{Pred. 10} \\
        \addlinespace[0.5mm]
        \multicolumn{3}{c}{\large{fRNN}} & \multicolumn{3}{c}{\large{RLadder}} & \multicolumn{3}{c}{\large{Srivastava \emph{et al.}}} \\
        \qim{0.265}{0.265}{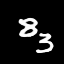} &
        \qim{0.265}{0.265}{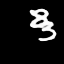} &
        \qim{0.265}{0.265}{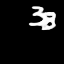} &
        \qim{0.265}{0.265}{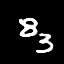} &
        \qim{0.265}{0.265}{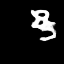} &
        \qim{0.265}{0.265}{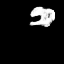} &
        \qim{0.265}{0.265}{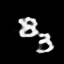} &
        \qim{0.265}{0.265}{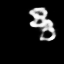} &
        \qim{0.265}{0.265}{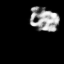} \\
        \qim{0.265}{0.20}{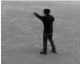} &
        \qim{0.265}{0.20}{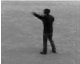} &
        \qim{0.265}{0.20}{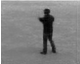} &
        \qim{0.265}{0.20}{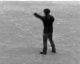} &
        \qim{0.265}{0.20}{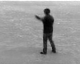} &
        \qim{0.265}{0.20}{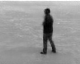} &
        \qim{0.265}{0.20}{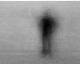} &
        \qim{0.265}{0.20}{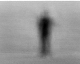} &
        \qim{0.265}{0.20}{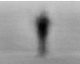} \\
        \qim{0.265}{0.20}{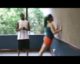} &
        \qim{0.265}{0.20}{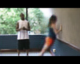} &
        \qim{0.265}{0.20}{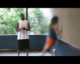} &
        \qim{0.265}{0.20}{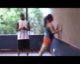} &
        \qim{0.265}{0.20}{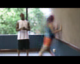} &
        \qim{0.265}{0.20}{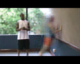} &
        \qim{0.265}{0.20}{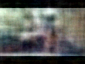} &
        \qim{0.265}{0.20}{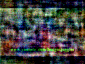} &
        \qim{0.265}{0.20}{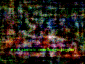} \\
        
        \multicolumn{3}{c}{\large{Mathieu \emph{et al.}}} & \multicolumn{3}{c}{\large{Villegas \emph{et al.}}} & \multicolumn{3}{c}{\large{Lotter \emph{et al.}}} \\
        \qim{0.265}{0.265}{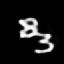} &
        \qim{0.265}{0.265}{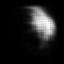} &
        \qim{0.265}{0.265}{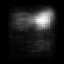} &
        \qim{0.265}{0.265}{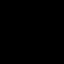} &
        \qim{0.265}{0.265}{images/qualitative/benchmark/mmnist/black} &
        \qim{0.265}{0.265}{images/qualitative/benchmark/mmnist/black} &
        \qim{0.265}{0.265}{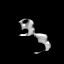} &
        \qim{0.265}{0.265}{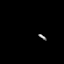} &
        \qim{0.265}{0.265}{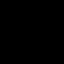} \\
        \qim{0.265}{0.20}{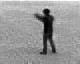} &
        \qim{0.265}{0.20}{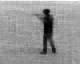} &
        \qim{0.265}{0.20}{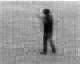} &
        \qim{0.265}{0.20}{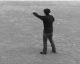} &
        \qim{0.265}{0.20}{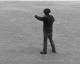} &
        \qim{0.265}{0.20}{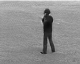} &
        \qim{0.265}{0.20}{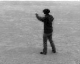} &
        \qim{0.265}{0.20}{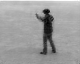} &
        \qim{0.265}{0.20}{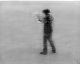} \\
        \qim{0.265}{0.20}{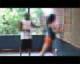} &
        \qim{0.265}{0.20}{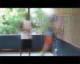} &
        \qim{0.265}{0.20}{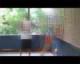} &
        \qim{0.265}{0.20}{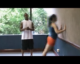} &
        \qim{0.265}{0.20}{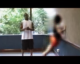} &
        \qim{0.265}{0.20}{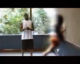} &
        \qim{0.265}{0.20}{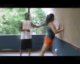} &
        \qim{0.265}{0.20}{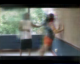} &
        \qim{0.265}{0.20}{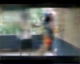} \\
    \end{tabular}}
    \vspace{-3mm}
    \caption{Predictions at 1, 5, and 10 time steps from the last ground truth frame. RLadder predictions on MMNIST are from the model pre-trained on KTH.}
    \label{fig:sidebyside}
    \vspace{-0.5cm}
\end{figure}

In Fig. \ref{fig:sidebyside} we compare predictions from the proposed approach against the RLadder baseline and other state of the art methods. For the MMNIST dataset we did not consider Villegas \emph{et al.} and Lotter \emph{et al.} since these methods fail to successfully converge and they predict a sequence of black frames. From the rest of approaches, fRNN obtains the best predictions, with little blur or distortion. The RLadder baseline is the second best approach. It does not introduce blur, but heavily deforms the digits after they cross. Srivastava \emph{et al.} and Mathieu \emph{et al.} both accumulate blur over time, but while the former does so to a smaller degree, the later makes the digits unrecognisable after five frames.

%This is probably due to the affine transform approach used by the later, allowing him to transform an image without encoding the appearance of the moving regions.

For KTH, Villegas \emph{et al.} obtains outstanding qualitative results. It predicts plausible dynamics and maintains the sharpness of both the individual and background. Both fRNN and RLadder follow closely, predicting plausible dynamics, but not being as good as Villegas \emph{et al.} at maintaining the sharpness of the individual. On UCF101 the best prediction is obtained by our model, with little blur or distortion compared to the other methods. The second best is Villegas \emph{et al.}, successfully capturing the movement patterns but introducing more blur and important distorsions on the last frame. When looking at the background, fRNN proposes a plausible initial estimation and progressively completes it as the woman moves. On the other hand, Villegas \emph{et al.} modifies already generated regions as more background is uncovered, generating an unrealistic sequence regarding the background. Srivastava and Lotter fail on both KTH and UCF101. Srivastava \emph{et al.} heavily distort the frames. As discussed in Section \ref{sec:experiments:quantitative}, this is due to the use of fully connected recurrent layers, which constrains the state size and prevents the model from encoding relevant information on more complex scenarios. In the case of Lotter, it makes good predictions for the first frame, but rapidly accumulates artefacts.

\subsection{Representation stratification analysis}
\label{sec:experiments:stratification}

Here we analyse the stratification of the sequence representation among the bGRU layers. Because bGRU units allow for a bijective mapping between states, it is possible to remove the deepest layers of a trained network, allowing us to check how the predictions are affected and providing an insight on the dynamics captured by each layer. Specifically, the same sequences are predicted multiple times, removing a layer each time. To our knowledge, this is the first topology allowing for a direct observation of the behaviour encoded on each layer.

\begin{figure}[t!]
    \centering
    \setlength{\tabcolsep}{0.25em}
    \resizebox{\columnwidth}{0.55\columnwidth}{\begin{tabular}{ccccccccccccccc}
        \includegraphics[]{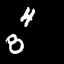} & 
        \includegraphics[]{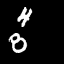} & 
        \includegraphics[]{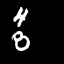} & 
        \includegraphics[]{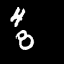} & 
        \includegraphics[]{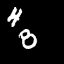} & 
        \includegraphics[]{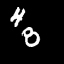} & 
        \includegraphics[]{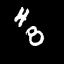} & 
        \includegraphics[]{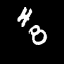} & 
        \includegraphics[]{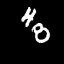} & 
        \includegraphics[]{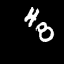} & 
        \includegraphics[]{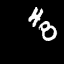} & 
        \includegraphics[]{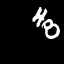} & 
        \includegraphics[]{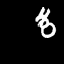} & 
        \includegraphics[]{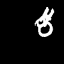} & 
        \includegraphics[]{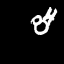} \\ 
        \includegraphics[]{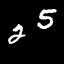} & 
        \includegraphics[]{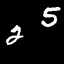} & 
        \includegraphics[]{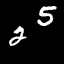} & 
        \includegraphics[]{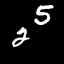} & 
        \includegraphics[]{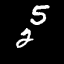} & 
        \includegraphics[]{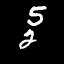} & 
        \includegraphics[]{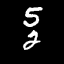} & 
        \includegraphics[]{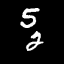} & 
        \includegraphics[]{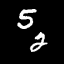} & 
        \includegraphics[]{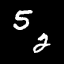} & 
        \includegraphics[]{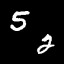} & 
        \includegraphics[]{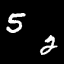} & 
        \includegraphics[]{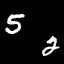} & 
        \includegraphics[]{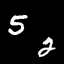} & 
        \includegraphics[]{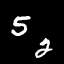} \\
        \rowcolor{gray!25}
        \multicolumn{5}{c}{\multirow{-2}{*}{{\Huge 8 bGRU layers}}} &
        \includegraphics[]{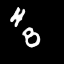} & 
        \includegraphics[]{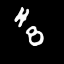} & 
        \includegraphics[]{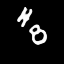} & 
        \includegraphics[]{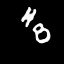} & 
        \includegraphics[]{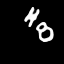} & 
        \includegraphics[]{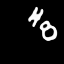} & 
        \includegraphics[]{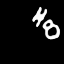} & 
        \includegraphics[]{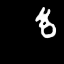} & 
        \includegraphics[]{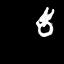} & 
        \includegraphics[]{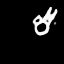} \\
        \rowcolor{gray!25}
        \multicolumn{5}{c}{} &
        \includegraphics[]{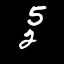} & 
        \includegraphics[]{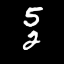} & 
        \includegraphics[]{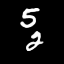} & 
        \includegraphics[]{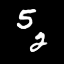} & 
        \includegraphics[]{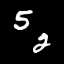} & 
        \includegraphics[]{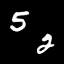} & 
        \includegraphics[]{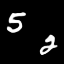} & 
        \includegraphics[]{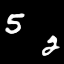} & 
        \includegraphics[]{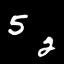} & 
        \includegraphics[]{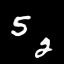} \\
        \multicolumn{5}{c}{\multirow{-2}{*}{{\Huge 6 bGRU layers}}} &
        \includegraphics[]{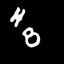} & 
        \includegraphics[]{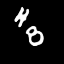} & 
        \includegraphics[]{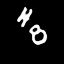} & 
        \includegraphics[]{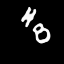} & 
        \includegraphics[]{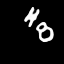} & 
        \includegraphics[]{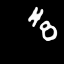} & 
        \includegraphics[]{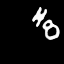} & 
        \includegraphics[]{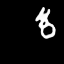} & 
        \includegraphics[]{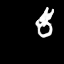} & 
        \includegraphics[]{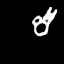} \\
        \multicolumn{5}{c}{} &
        \includegraphics[]{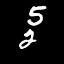} & 
        \includegraphics[]{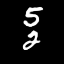} & 
        \includegraphics[]{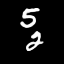} & 
        \includegraphics[]{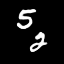} & 
        \includegraphics[]{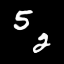} & 
        \includegraphics[]{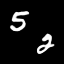} & 
        \includegraphics[]{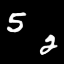} & 
        \includegraphics[]{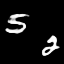} & 
        \includegraphics[]{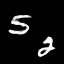} & 
        \includegraphics[]{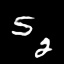} \\
        \rowcolor{gray!25}
        \multicolumn{5}{c}{\multirow{-2}{*}{{\Huge 5 bGRU layers}}} &
        \includegraphics[]{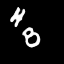} & 
        \includegraphics[]{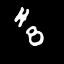} & 
        \includegraphics[]{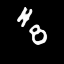} & 
        \includegraphics[]{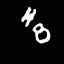} & 
        \includegraphics[]{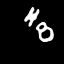} & 
        \includegraphics[]{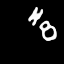} & 
        \includegraphics[]{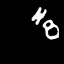} & 
        \includegraphics[]{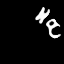} & 
        \includegraphics[]{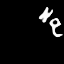} & 
        \includegraphics[]{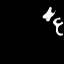} \\
        \rowcolor{gray!25}
        \multicolumn{5}{c}{} &
        \includegraphics[]{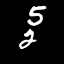} & 
        \includegraphics[]{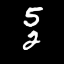} & 
        \includegraphics[]{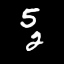} & 
        \includegraphics[]{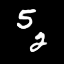} & 
        \includegraphics[]{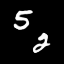} & 
        \includegraphics[]{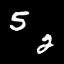} & 
        \includegraphics[]{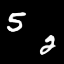} & 
        \includegraphics[]{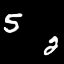} & 
        \includegraphics[]{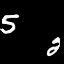} & 
        \includegraphics[]{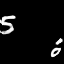} \\
        \multicolumn{5}{c}{\multirow{-2}{*}{{\Huge 4 bGRU layers}}} &
        \includegraphics[]{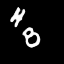} & 
        \includegraphics[]{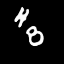} & 
        \includegraphics[]{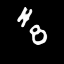} & 
        \includegraphics[]{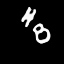} & 
        \includegraphics[]{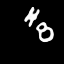} & 
        \includegraphics[]{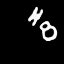} & 
        \includegraphics[]{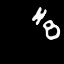} & 
        \includegraphics[]{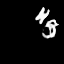} & 
        \includegraphics[]{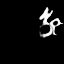} & 
        \includegraphics[]{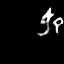} \\
        \multicolumn{5}{c}{} &
        \includegraphics[]{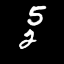} & 
        \includegraphics[]{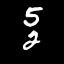} & 
        \includegraphics[]{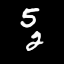} & 
        \includegraphics[]{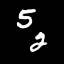} & 
        \includegraphics[]{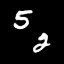} & 
        \includegraphics[]{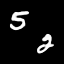} & 
        \includegraphics[]{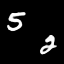} & 
        \includegraphics[]{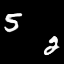} & 
        \includegraphics[]{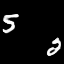} & 
        \includegraphics[]{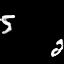} \\
        \rowcolor{gray!25}
        \multicolumn{5}{c}{\multirow{-2}{*}{{\Huge 3 bGRU layers}}} &
        \includegraphics[]{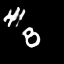} & 
        \includegraphics[]{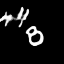} & 
        \includegraphics[]{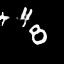} & 
        \includegraphics[]{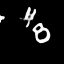} & 
        \includegraphics[]{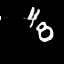} & 
        \includegraphics[]{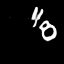} & 
        \includegraphics[]{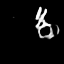} & 
        \includegraphics[]{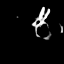} & 
        \includegraphics[]{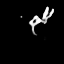} & 
        \includegraphics[]{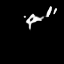} \\
        \rowcolor{gray!25}
        \multicolumn{5}{c}{} &
        \includegraphics[]{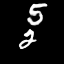} & 
        \includegraphics[]{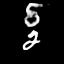} & 
        \includegraphics[]{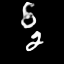} & 
        \includegraphics[]{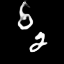} & 
        \includegraphics[]{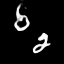} & 
        \includegraphics[]{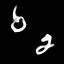} & 
        \includegraphics[]{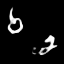} & 
        \includegraphics[]{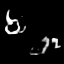} & 
        \includegraphics[]{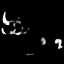} & 
        \includegraphics[]{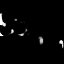} \\
        \multicolumn{5}{c}{\multirow{-2}{*}{{\Huge 2 bGRU layers}}} &
        \includegraphics[]{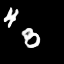} & 
        \includegraphics[]{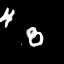} & 
        \includegraphics[]{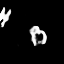} & 
        \includegraphics[]{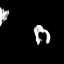} & 
        \includegraphics[]{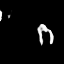} & 
        \includegraphics[]{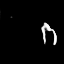} & 
        \includegraphics[]{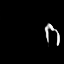} & 
        \includegraphics[]{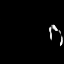} & 
        \includegraphics[]{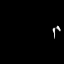} & 
        \includegraphics[]{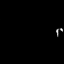} \\
        \multicolumn{5}{c}{} &
        \includegraphics[]{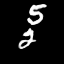} & 
        \includegraphics[]{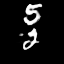} & 
        \includegraphics[]{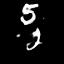} & 
        \includegraphics[]{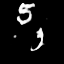} & 
        \includegraphics[]{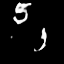} & 
        \includegraphics[]{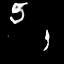} & 
        \includegraphics[]{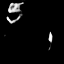} & 
        \includegraphics[]{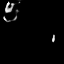} & 
        \includegraphics[]{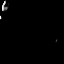} & 
        \includegraphics[]{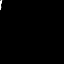} \\
        \rowcolor{gray!25}
        \multicolumn{5}{c}{\multirow{-2}{*}{{\Huge 1 bGRU layer}}} &
        \includegraphics[]{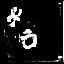} & 
        \includegraphics[]{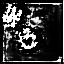} & 
        \includegraphics[]{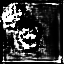} & 
        \includegraphics[]{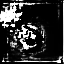} & 
        \includegraphics[]{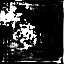} & 
        \includegraphics[]{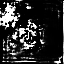} & 
        \includegraphics[]{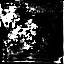} & 
        \includegraphics[]{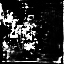} & 
        \includegraphics[]{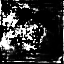} & 
        \includegraphics[]{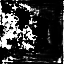} \\
        \rowcolor{gray!25}
        \multicolumn{5}{c}{} &
        \includegraphics[]{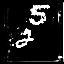} & 
        \includegraphics[]{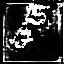} & 
        \includegraphics[]{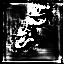} & 
        \includegraphics[]{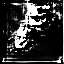} & 
        \includegraphics[]{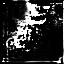} & 
        \includegraphics[]{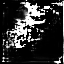} & 
        \includegraphics[]{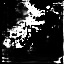} & 
        \includegraphics[]{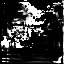} & 
        \includegraphics[]{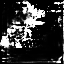} & 
        \includegraphics[]{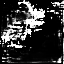} \\
        \multicolumn{5}{c}{\multirow{-2}{*}{{\Huge 0 bGRU layers}}} &
        \includegraphics[]{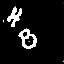} & 
        \includegraphics[]{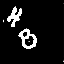} & 
        \includegraphics[]{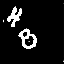} & 
        \includegraphics[]{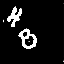} & 
        \includegraphics[]{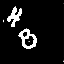} & 
        \includegraphics[]{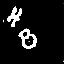} & 
        \includegraphics[]{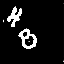} & 
        \includegraphics[]{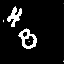} & 
        \includegraphics[]{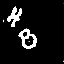} & 
        \includegraphics[]{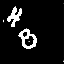} \\
        \multicolumn{5}{c}{} &
        \includegraphics[]{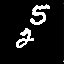} & 
        \includegraphics[]{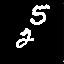} & 
        \includegraphics[]{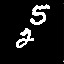} & 
        \includegraphics[]{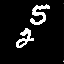} & 
        \includegraphics[]{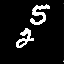} & 
        \includegraphics[]{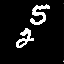} & 
        \includegraphics[]{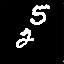} & 
        \includegraphics[]{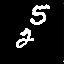} & 
        \includegraphics[]{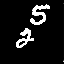} & 
        \includegraphics[]{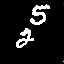} \\
    \end{tabular}}
    \vspace{-0.4cm}
    \caption{\small Moving MNIST predictions with fRNN layer removal. Removing all bGRU layers (last row) leaves two convolutional layers and their transposed convolutions, providing an identity mapping.}
    \label{fig:stratified:mmnist}\vspace{-0.4cm}
\end{figure}

This is shown in Fig. \ref{fig:stratified:mmnist} for the MMNIST dataset. The analysed model consists of 10 layers: 2 convolutional layers and 8 bGRU layers. Firstly, removing the last 2 bGRU layers has no significant impact on prediction. This shows that, for this simple dataset, the network has a higher capacity than required. Further removing layers does not result in a loss of pixel-level information, but on a progressive loss of behaviours, from more complex to simpler ones. This means information at a given level of abstraction is not encoded into higher level layers. When removing the third deepest bGRU layer, the digits stop bouncing and keep their linear trajectories, exiting the image. This indicates this layer is in charge of encoding information on bouncing dynamics. When removing the next layer, digits stop behaving correctly on the boundaries of the image. Parts of the digit bounce while others keep the previous trajectory. While this also has to do with bouncing dynamics, the layer seems to be in charge of recognising digits as single units following the same movement pattern. When removed, different segments of the digit are allowed to move as separate elements. Finally, with only 3-2 bGRU layers the digits are distorted in various ways. With only two layers left, the general linear dynamics are still captured by the model. By leaving a single bGRU layer, the linear dynamics are lost.

According to these results, linear movement dynamics are captured at pixel level on the first two bGRU layers. The next two start aggregating these movement patterns into single-trajectory components, preventing their distortion. The collision of these components with image bounds is also detected. The fifth layer aggregates single-motion components into digits, forcing them to follow the same motion. This seems to have the effect of preventing bounces, likely due to only one of the components reaching the edge of the image. It is the sixth bGRU layer that provides a coherent bouncing pattern for the whole digit.

% --------------------------------------------------------
% --------------------------------------------------------
\section{Conclusions}
\label{sec:conclusions}

We presented Folded Recurrent Neural Networks, a new recurrent architecture for video prediction with lower computational and memory cost compared to equivalent recurrent AE models. This is achieved by using the proposed bijective GRUs, which horizontally pass information between the encoder and decoder. This eliminates the need for using the entire AE at any given step: only the encoder or decoder needs to be executed for both input encoding and prediction, respectively. It also facilitates the convergence by naturally providing a noisy identity function during training. We evaluated our approach on three video datasets, outperforming state of the art prediction results on MMNIST and UCF101, and obtaining competitive results on KTH with 2 and 3 times less memory usage and computational cost than the best scored approach. Qualitatively, the model can limit and recover from blur by preventing its propagation from low to high level dynamics. We also demonstrated stratification of the representation, topology optimisation, and model explainability through layer removal. %This is a straightforward process in this topology. 
Each layer has been shown to modify the state of the previous one by adding more complex behaviours: removing a layer eliminates its behaviours but leaves lower-level ones untouched.

\clearpage

\bibliographystyle{splncs}
\bibliography{main.bib}
\end{document}